\documentclass[letterpaper, 10 pt, conference]{ieeeconf}  

\IEEEoverridecommandlockouts                              

\usepackage[]{graphicx}    
\usepackage[colorlinks=true, urlcolor=blue, linkcolor=black, citecolor=black]{hyperref} 
\newcommand{\ignore}[1]{}  
\usepackage{float}
\usepackage{amsmath} 
\usepackage[compatibility=false]{caption}
\usepackage{subcaption}
\makeatletter
\let\labelindent\relax
\makeatother
\usepackage{enumitem}
\setlist[enumerate]{leftmargin=*}

\title{\LARGE \bf
Neural Aided Kalman Filtering for UAV State Estimation in Degraded Sensing Environments
}

\author{Akhil Gupta $^{1}$ and Erhan Guven$^{1}$
\thanks{\footnotesize 979-8-3315-7360-7/26/\$31.00 \copyright2026 IEEE}
\thanks{$^{1}$Akhil Gupta and Erhan Guven are with the Whiting School of Engineering EP Program, Johns Hopkins University, Baltimore, Maryland, USA.
{\tt\small Akhil Gupta, Erhan Guven}}%
}

\begin{document}

\maketitle
\thispagestyle{empty}
\pagestyle{empty}

\begin{abstract}

Accurate state estimation of nonlinear dynamical systems is fundamental to modern aerospace operations across the air, sea, and space domains. Reliable tracking and control require precise knowledge of a platform’s position and velocity state to predict near-term trajectories. Online tracking of adversarial remote-controlled unmanned aerial vehicles (UAVs) can be especially challenging due to agile nonlinear motion, noisy and sparse sensor measurement availability, and unknown control inputs, which violate key assumptions of classical Kalman filter variants and can degrade overall performance. These limitations motivate alternative approaches that capture control-induced nonlinear dynamics while remaining suitable for real-time deployment. Neural networks (NNs) can learn complex nonlinear relationships from data, but their lack of principled uncertainty quantification limits use in state estimation tasks where confidence bounds are critical. We address this using Bayesian Neural Networks (BNNs), which model uncertainty through distributions over network weights and produce predictive means and uncertainties via Monte Carlo sampling. Building on this capability, we propose the Bayesian Neural Kalman Filter (\textit{BNKF}), a hybrid framework that couples a trained BNN with a Kalman filter correction step for robust online UAV state estimation. Unlike related neural Kalman approaches, \textit{BNKF} produces full state predictions and incorporates Bayesian uncertainty directly into covariance propagation, improving robustness in high-noise regimes. We evaluate \textit{BNKF} under varying radar noise levels and sampling rates using synthetic nonlinear UAV flight data. Five-fold cross-validation shows that \textit{BNKF} outperforms Extended and Unscented Kalman Filters in accuracy, precision, and state containment under higher noise conditions. An ensemble variant (\textit{BNKFe}) further improves precision in high-noise edge cases at the slight expense of accuracy, while overall runtime analysis confirms minimal inference overhead and the potential for real-time deployment feasibility.

\end{abstract}

\section{Introduction}
Accurate state estimation of nonlinear dynamical systems is essential in aerospace engineering and autonomous systems \cite{electronics13112208}. Designing effective tracking and control systems requires reliable knowledge of an object’s current state and near-term trajectory \cite{AVZAYESH2021102912}. Real-world estimation tasks introduce additional complexities, including noisy sensor measurements, missing data, and limited or ambiguous information within individual observations. In this work, we define state estimation as the task of approximating an object's position over time based on historical and incoming sensor measurements, known sensor uncertainty characteristics, and prior knowledge of the physical dynamics governing the object’s motion. Traditional approaches such as the Kalman Filter (KF) and its nonlinear variants have demonstrated strong performance across this type of state estimation task \cite{electronics13112208}, but can degrade quickly under high measurement noise or infrequent observations—particularly when the system dynamics and control inputs are difficult to model \cite{kf_shortcoming_cov}. 
    
Unmanned aerial vehicles (UAVs), which are expected to see substantial growth due to their low cost and maneuverability \cite{Drone_Growth}, present a particularly challenging case. Unlike conventional aircraft, UAV motion can be driven by control policies that are difficult to model explicitly. Their agility and broad range of control behaviors introduce significant nonlinearity that may not be captured adequately by classical KF formulations, which often rely on simplified or linearized dynamics \cite{1642720}.

Data-driven neural networks can learn highly nonlinear relationships from historical data \cite{NN_Value}, but standard deterministic networks provide limited uncertainty quantification, which is critical for estimation and decision-making. To address this, we leverage Bayesian Neural Networks (BNNs), which represent uncertainty through distributions over network weights and produce predictive uncertainty via Monte Carlo sampling. We use a trained variational BNN as a surrogate for the prediction component of a Kalman-style estimator, and then apply a standard correction step when new measurements arrive. We refer to this hybrid approach as the Bayesian Neural Kalman Filter (\textit{BNKF}). Our goal is to evaluate whether an \emph{offline-learned motion prior} combined with a probabilistic correction step improves robustness under \emph{degraded sensing conditions} (e.g., increased measurement noise and reduced sampling).

Altogether this paper makes three contributions:
(1) We introduce \textit{BNKF}, a hybrid estimator that couples a variational BNN with a Kalman correction step, using BNN predictive uncertainty as an explicit input to uncertainty handling during online estimation. (2) We evaluate \textit{BNKF} under controlled measurement degradation by varying radar measurement noise levels and sampling rates on nonlinear UAV trajectories, and compare against EKF and UKF baselines. (3) We analyze estimation accuracy, uncertainty, and truth containment across noise regimes, and explicitly discuss limitations of simulation-only evaluation and Gaussian noise assumptions.

The remainder of this paper is organized as follows. Section~\ref{sec:related_work} reviews related work. Section~\ref{sec:problem_modeling} introduces the problem formulation and data generation. Section~\ref{sec:algorithms} details the implementation of EKF, UKF, \textit{BNKF}, and \textit{BNKFe}. Section~\ref{sec:evaluation} describes the evaluation methodology and metrics, and Section~\ref{sec:results} presents results and discussion.
\section{Related Work}\label{sec:related_work}

Several lines of prior research intersect with the topic of this paper. For instance, deep learning methods have been explored as replacements for traditional linear solvers, particularly in applications such as solving the Poisson equation \cite{Markidis2021TheOA}. Physics-Informed Neural Networks (PINNs) have shown promise in modeling nonlinear dynamics, including recent work focused on tracking newly formed space debris following inelastic collisions in Low-Earth Orbit \cite{space-debris-tracking} and UAV state estimation \cite{electronics13112208}. DeepKF employs three LSTMs as a surrogate for the transition function in the Kalman filter’s prediction step, whereas KalmanNet uses a compact neural network to dynamically estimate the Kalman gain during the update step \cite{kalmannet} \cite{lstm-kf}.

In contrast to these prior neural–Kalman approaches we focus specifically on leveraging variational Bayesian neural networks (BNNs) as predictive surrogates that output both a state estimate and an associated predictive uncertainty \cite{bnn_uncer_explain}. Rather than trying to solely learn residual corrections or gain mappings, our formulation explicitly carries Monte Carlo–derived predictive uncertainty from a BNN into the subsequent Kalman correction step. DeepKF differs in the sense that it prefers a historical set of prior sensor measurements whereas \textit{BNKF} only requires the first preceding. We do not claim to be the first to combine neural networks with Kalman filtering; instead, our contribution lies in examining how data-driven Bayesian predictive uncertainty can be integrated into covariance handling under degraded sensing conditions, and in systematically evaluating this hybrid formulation across controlled noise and sampling regimes relative to EKF and UKF baselines.

\section{Problem Modeling}\label{sec:problem_modeling}
\begin{figure}[t] 
    \centering
    \begin{subfigure}{0.32\columnwidth}
        \centering
        \includegraphics[width=\linewidth]{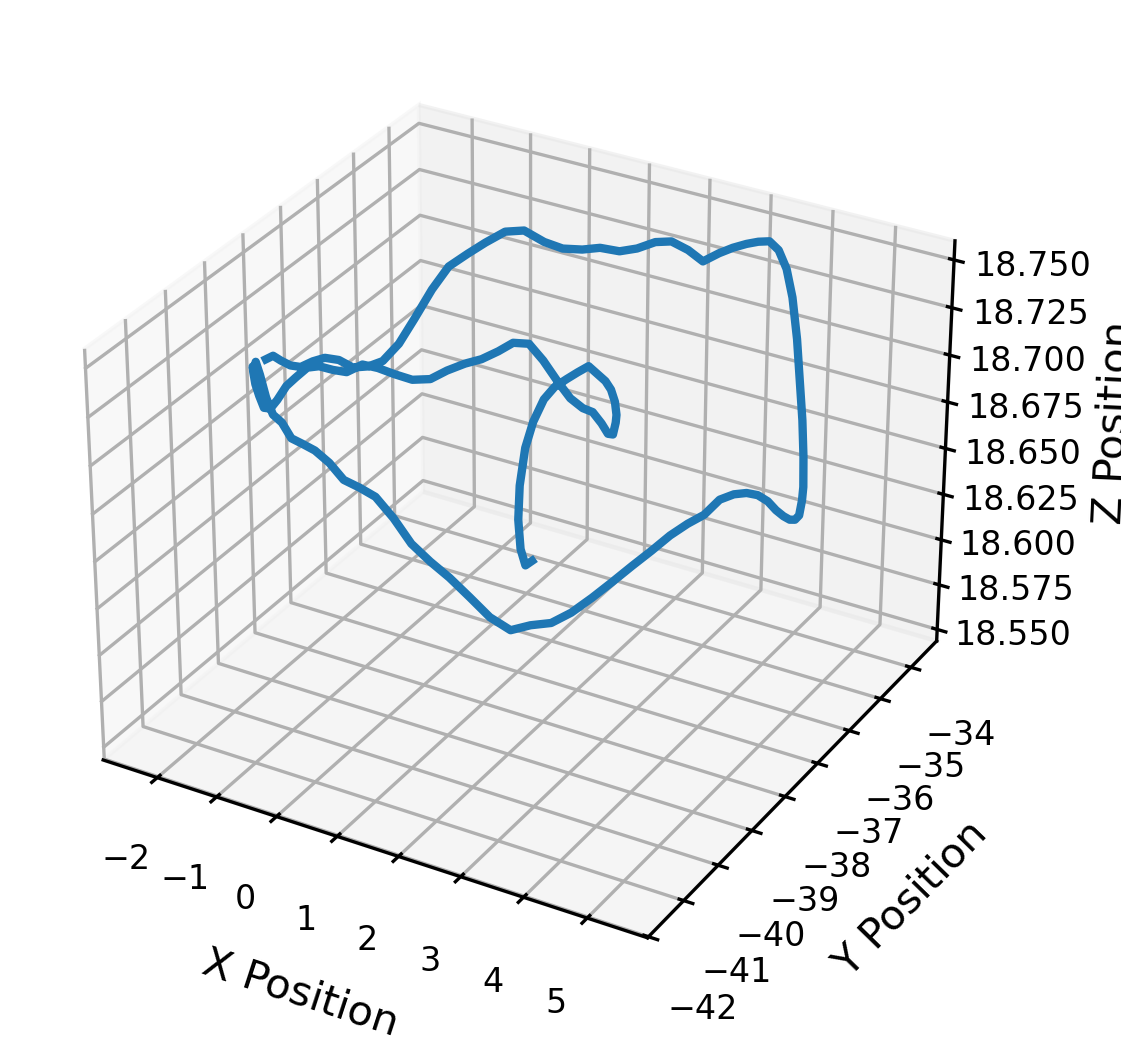}
        \caption{Trajectory 1}
        \label{fig:traj1}
    \end{subfigure}
    \hfill
    \begin{subfigure}{0.32\columnwidth}
        \centering
        \includegraphics[width=\linewidth]{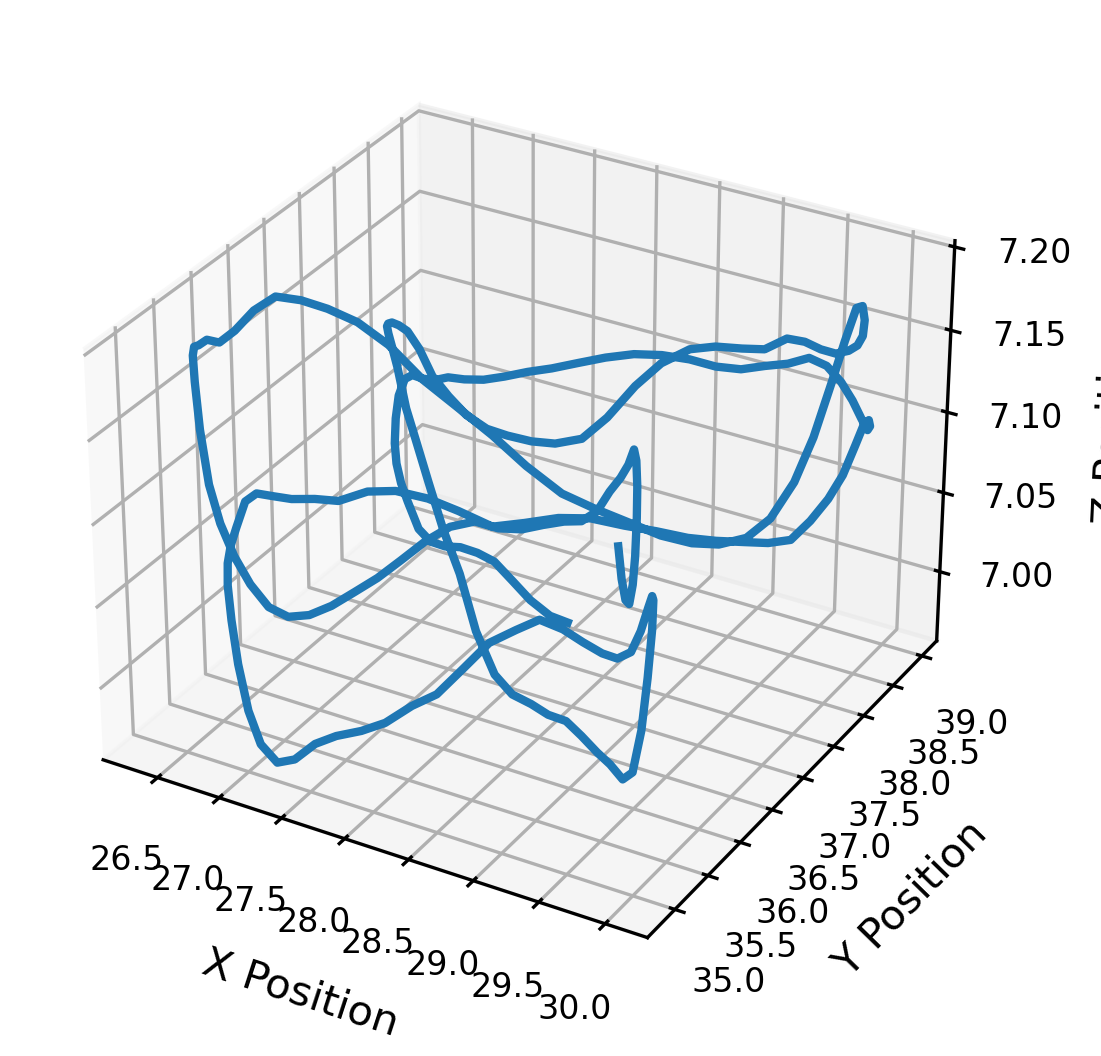}
        \caption{Trajectory 2}
        \label{fig:traj2}
    \end{subfigure}
    \hfill
    \begin{subfigure}{0.32\columnwidth}
        \centering
        \includegraphics[width=\linewidth]{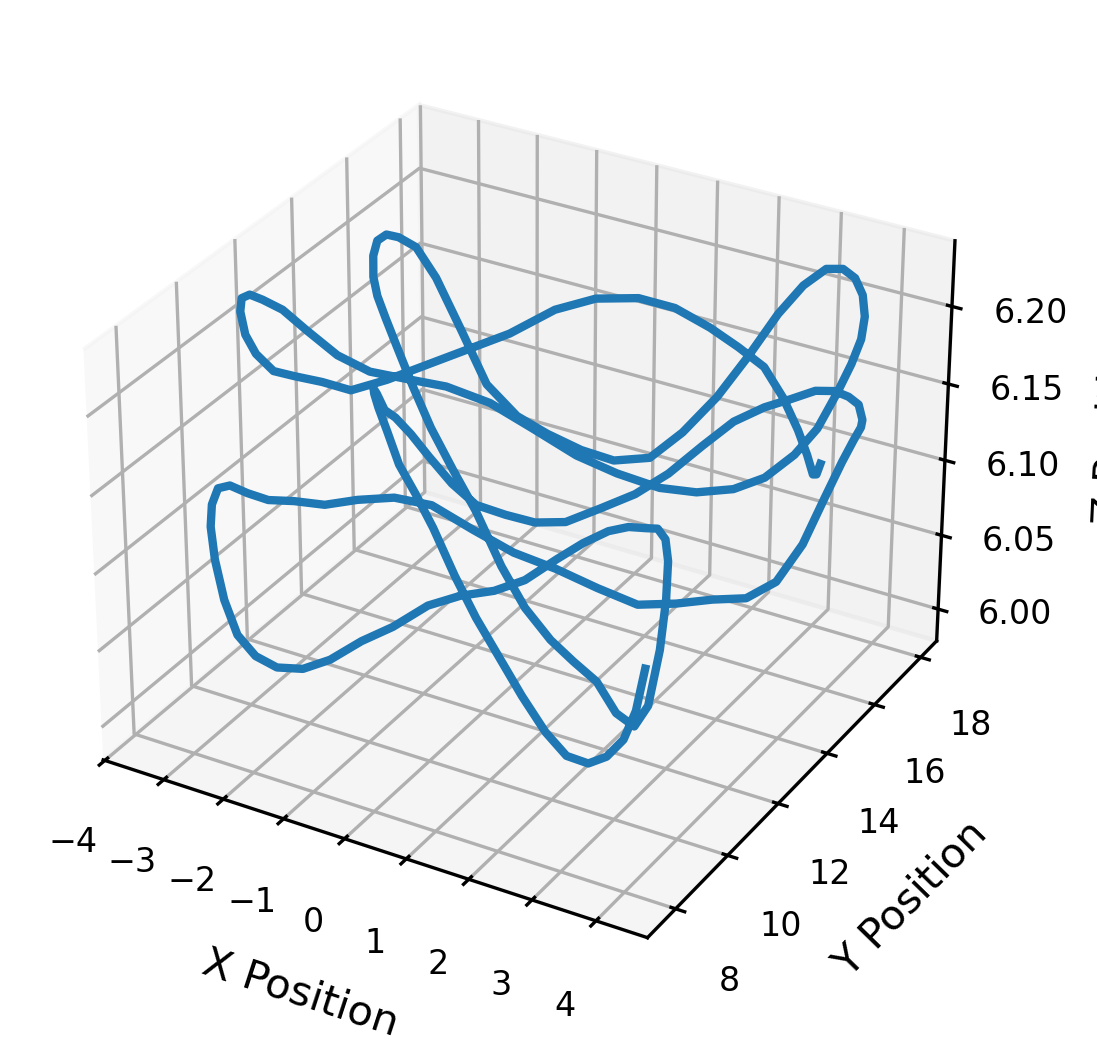}
        \caption{Trajectory 3}
        \label{fig:traj3}
    \end{subfigure}
    \caption{Example Simulated Trajectories.}
    \label{fig:three_trajs}
\end{figure}

In this section, we describe our approach to modeling the general state estimation problem. We utilize the \textit{Synthetic-UAV-Flight-Trajectories} dataset available on Hugging Face, which contains over 5,000 randomized UAV flight trajectories spanning approximately 20 hours of simulated flight time \cite{drone-dataset}. The data is generated using the Gazebo simulator, which leverages a high-fidelity physics engine to model realistic UAV dynamics and environmental interactions \cite{UAV_Dataset_Paper}. This dataset was designed to support research and experimentation in data-driven trajectory prediction and control for aerial platforms \cite{drone-dataset}. As illustrated in Figure~\ref{fig:three_trajs}, the trajectories exhibit nonlinear motion patterns, making them particularly suitable for evaluating nonlinear state estimation techniques.

Within each trajectory, we are provided with 3D position and velocity points in the XYZ frame with meter units, along with a timestamp that corresponds to each point. For the purposes of modeling our study, we consider these data points as the \textbf{true} state of the drone at any time. Consequently, we use \(\mathbf{x}_{t}\) to refer to the true and full position state of the drone at time \textit{t} (Equation~\ref{eq:true_position_state}).

The true position state at time $t$ is defined as
\begin{equation}
    \mathbf{x}_{t} =
    \begin{bmatrix}
        x_{t} & y_{t} & z_{t}
    \end{bmatrix}
    \label{eq:true_position_state}
\end{equation}

\begin{figure}[t] 
    \centering
    \includegraphics[width=\columnwidth]{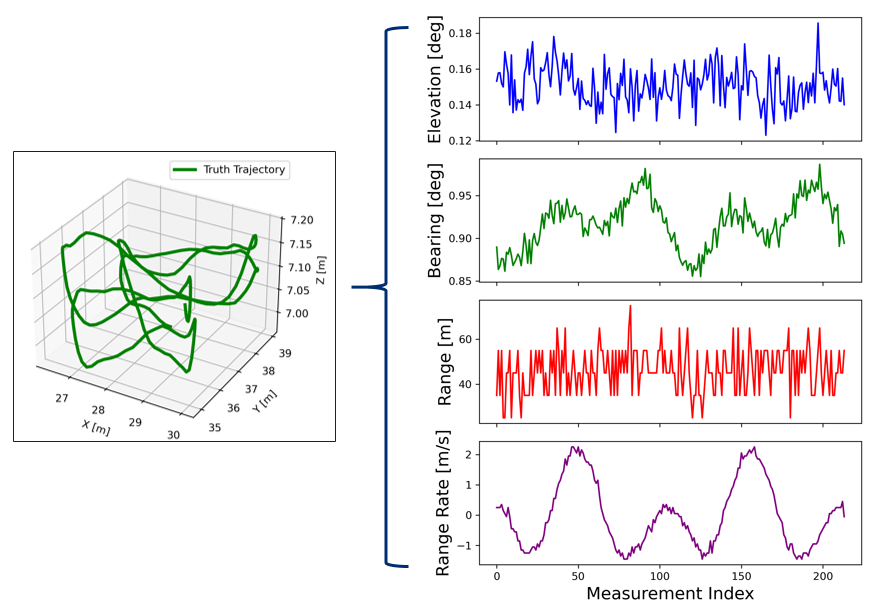}
    \caption{Radar Sensor Measurement Generation Example}
    \label{fig:radar_sens_gen}
\end{figure}

To simulate the types of measurements typically received from a real-world radar system, we employed the Stone Soup library \cite{stonesoup2020} in Python. This tool was used to generate noisy observations—specifically, range, range rate, bearing, and elevation—for each UAV trajectory based on a fixed sensor location (Figure~\ref{fig:radar_sens_gen}). These measurements were computed using the true position and velocity states derived earlier with added noise. To evaluate the robustness of our approach under diverse sensing conditions, we systematically varied two key parameters during simulation: the sensor noise level and the measurement sampling rate.  
In our context, noise level refers to the magnitude of the sigma value used to generate each respective measurement type. We had three discrete bins of noise levels: low, medium, and high - for which the actual values are defined in Table~\ref{tab:sigma_levels}.

\begin{table}[H]
\renewcommand{\arraystretch}{1.5}
\centering
\resizebox{\columnwidth}{!}{%
\begin{tabular}{|c|c|c|c|c|}
\hline
\textbf{Level} & \textbf{Range [m]} & \textbf{Range-Rate [m/s]} & \textbf{Bearing [deg]} & \textbf{Elevation [deg]} \\
\hline 
Low    & 1   & 0.01 & 0.001 & 0.001 \\
\hline
Medium & 10  & 0.1  & 0.01  & 0.01  \\
\hline
High   & 100 & 1    & 0.1   & 0.1   \\
\hline
\end{tabular}
}
\caption{$1\sigma$ Measurement Uncertainty at Different Levels}
\label{tab:sigma_levels}
\end{table}

\begin{figure}[t] 
    \centering
    \begin{subfigure}{0.32\columnwidth}
        \centering
        \includegraphics[width=\linewidth]{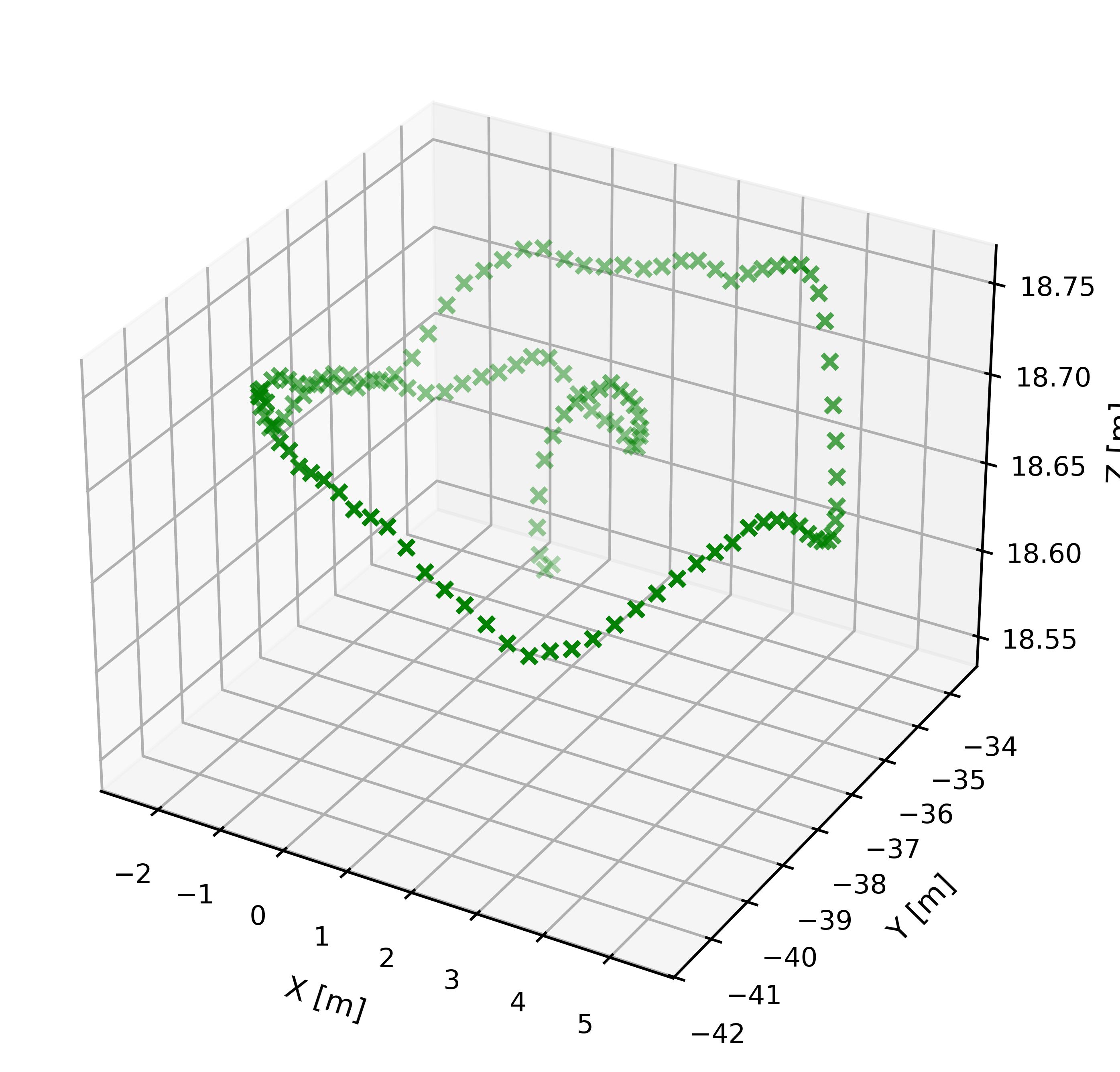}
        \caption{1.0 SR}
        \label{fig:traj1_sr0}
    \end{subfigure}
    \hfill
    \begin{subfigure}{0.32\columnwidth}
        \centering
        \includegraphics[width=\linewidth]{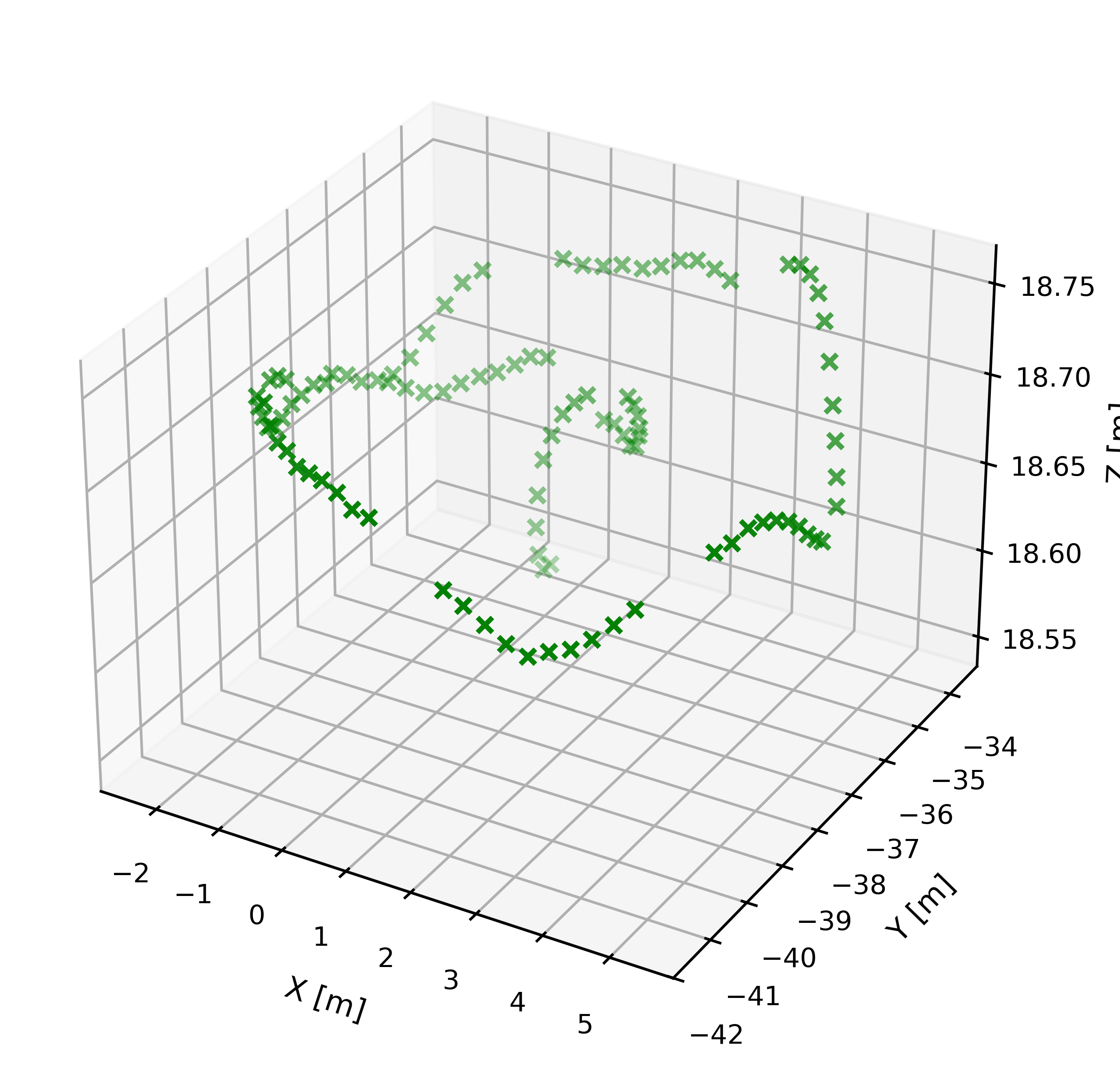}
        \caption{0.75 SR}
        \label{fig:traj2_sr23}
    \end{subfigure}
    \hfill
    \begin{subfigure}{0.32\columnwidth}
        \centering
        \includegraphics[width=\linewidth]{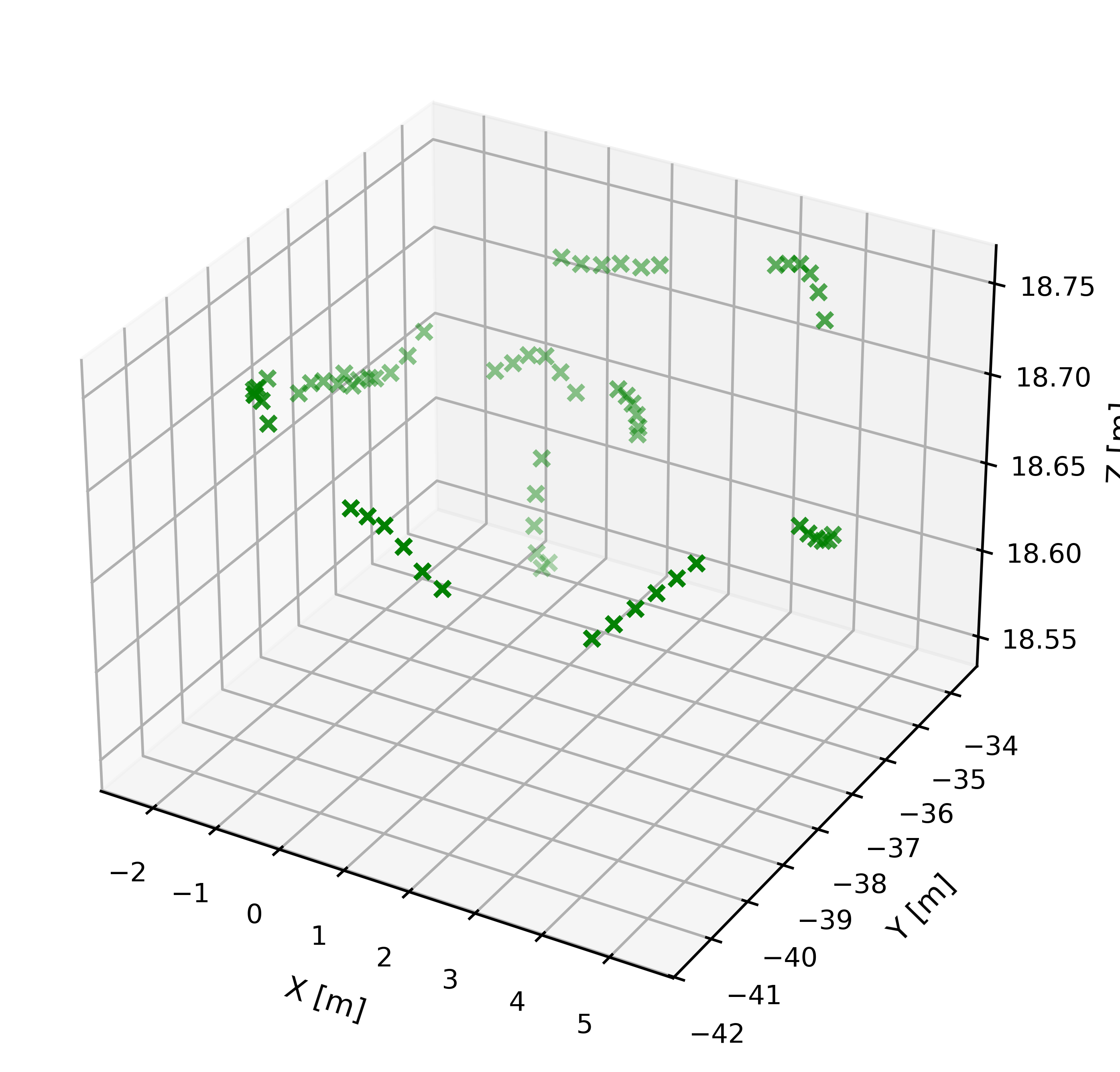}
        \caption{0.50 SR}
        \label{fig:traj3_50}
    \end{subfigure}
    \caption{Example Downsampled Trajectories}
    \label{fig:sample_removed_trajectories}
\end{figure}

The sampling rate refers to the proportion of measurements retained in a given trajectory. For instance, a sampling rate of 0.75 indicates that 75 percent of the original measurements are preserved, while 25 percent are omitted. This setup reflects real-world conditions, where a radar sensor may not always operate at full capacity due to limitations induced by its mission requirements or external interference. To augment the dataset and assess performance under reduced observation conditions, we generated additional versions of each trajectory with sampling rates of 0.75 and 0.50 (Figure~\ref{fig:sample_removed_trajectories}), alongside the original full-measurement sequences. These downsampled trajectories were concatenated into the train/test datasets for both the baseline and proposed models. Our hypothesis is that, under lower sampling conditions, the benchmark Kalman Filter's position estimates will diverge more frequently and with greater magnitude from the true state.

While each model approach for our experiments is distinct in nature, we ensure that the information content of the input feature vectors is consistent across models, and that the resulting output at a given time step remains equivalent for each algorithm 
$f$ (Equation~\ref{eq:general_state_function}) to promote fair evaluation. By accounting for this we are able to evaluate the \textit{BNKF} against the baseline algorithms in a consistent and unbiased manner. 

The feature vector at time $t$ is defined as
{
\setlength{\arraycolsep}{1pt}
\begin{equation}
\mathbf{z}_{t} =
\left[
\begin{array}{cccccccccccc}
\tilde{\rho}_{t} &
\tilde{\theta}_{t} &
\tilde{\phi}_{t} &
\tilde{\dot{\rho}}_{t} &
\tilde{\rho}_{t+1} &
\tilde{\theta}_{t+1} &
\tilde{\phi}_{t+1} &
\tilde{\dot{\rho}}_{t+1} &
\sigma_{\rho} &
\sigma_{\theta} &
\sigma_{\phi} &
\sigma_{\dot{\rho}}
\end{array}
\right]
\label{eq:general_feature_vector}
\end{equation}
}
where $\tilde{\rho}_{t}$, $\tilde{\theta}_{t}$, $\tilde{\phi}_{t}$, and $\tilde{\dot{\rho}}_{t}$ denote the noisy range, elevation, azimuth, and range-rate measurements at time $t$, respectively, and $\sigma_{\rho}$, $\sigma_{\theta}$, $\sigma_{\phi}$, and $\sigma_{\dot{\rho}}$ are the corresponding measurement noise standard deviations.

The estimated position state at time $t$ is given by
\begin{equation}
    \hat{\mathbf{x}}_{t}, \; \mathbf{P}_{t} = f(\mathbf{z}_{t}),
    \label{eq:general_state_function}
\end{equation}
where $f$ denotes a general state estimation method applied to the observations $\mathbf{z}_{t}$, with $\hat{x}_{t}$ representing the estimated state and $\mathbf{P}_{t}$ the associated uncertainty (e.g., covariance).

Moving forward we evaluate how well \textit{BNKF} reduces error between the estimated state $\hat{\mathbf{x}}_{t}$ and true state $\mathbf{x}_{t}$ from the feature vector $\mathbf{z}_{t}$ under varying Gaussian sensor noise, relative to EKF and UKF baselines. We additionally assess predictive uncertainty and truth containment.

\section{Algorithm Implementations}\label{sec:algorithms}
This section provides an overview of the different algorithms used in this comparison study -  including the proposed \textit{BNKF}. To support our evaluation, we implement four different algorithms described below.

\subsection{Extended Kalman Filter}
Because converting range, range-rate, bearing, and elevation measurements to Cartesian coordinates introduces nonlinear transformations, the extended Kalman filter (EKF) is required. The EKF is a widely used method for state estimation in nonlinear dynamical systems \cite{article_ekf}, extending the standard Kalman filter through local linearization about the current mean and covariance estimate \cite{4378854}. 

To establish a baseline for our study, we implement a standard EKF that combines a state-space and observation model to perform prediction and correction using noisy measurements. Although computationally efficient, the EKF can become unstable in highly nonlinear or noisy systems due to errors introduced by discrete linearization \cite{1642720}. We therefore use the EKF as a baseline across varying sensor uncertainty scenarios, hypothesizing that the \textit{BNKF} will outperform it in state accuracy under higher noise and data sparsity conditions. The EKF formulation used in this work is summarized below through its prediction and correction steps:

\textbf{Prediction Step:}  
\begin{equation}
    \hat{\mathbf{x}}_{t|t-1} = f(\hat{\mathbf{x}}_{t-1|t-1}, \mathbf{u}_{t-1})
    \label{eq:ekf_predict_state}
\end{equation}
\begin{equation}
    \mathbf{P}_{t|t-1} = \mathbf{F}_{t-1}\mathbf{P}_{t-1|t-1}\mathbf{F}_{t-1}^\top + \mathbf{Q}_{t-1}
    \label{eq:ekf_predict_uncer}
\end{equation}
where $f(\cdot)$ is the nonlinear transition, $\mathbf{F}_{t-1}$ its Jacobian, and $\mathbf{Q}_{t-1}$ the process noise covariance.

\textbf{Correction Step:}  
\begin{equation}
    \hat{\mathbf{x}}_{t} = \hat{\mathbf{x}}_{t|t-1} + \mathbf{K}_t\!\left(\mathbf{z}_t - h(\hat{\mathbf{x}}_{t|t-1})\right)
    \label{eq:ekf_update_state}
\end{equation}
\begin{equation}
    \mathbf{P}_{t} = (\mathbf{I} - \mathbf{K}_t\mathbf{H}_t)\mathbf{P}_{t|t-1}
    \label{eq:ekf_update_uncer}
\end{equation}
where $h(\cdot)$ is the measurement function, $\mathbf{H}_t$ its Jacobian, and $\mathbf{K}_t$ the Kalman gain.

The EKF produces a final state estimate $\hat{\mathbf{x}}_{t}$ and covariance $\mathbf{P}_{t}$. In practice, we use the Stone Soup Python library for implementation \cite{stonesoup2020}.

\subsection{Unscented Kalman Filter (UKF)}
The UKF performs nonlinear state estimation without explicit linearization by propagating a set of sigma points through the system dynamics and measurement models to approximate the mean and covariance. It is often considered superior to the EKF for estimating nonlinear trajectories \cite{ukf_superior_ekf}, making it a natural benchmark for strongly nonlinear scenarios such as ours. However, UKF performance can degrade under high measurement noise or when noise characteristics are poorly modeled, leading to increased uncertainty and reduced estimation accuracy \cite{ukf_high_noise_shortcomings}. The filter may also diverge under sparse measurement conditions when the underlying control law is unknown in UAV tracking scenarios. Although many UKF variants exist, we benchmark the original formulation by Julier et al. \cite{original_ukf_paper}. The UKF follows the standard Kalman filtering cycle of prediction and correction using past and incoming measurements. While the process is described conceptually here, our implementation uses the Python Stone Soup library for computational efficiency \cite{stonesoup2020}.

\textbf{Prediction Step:}  
\begin{equation}
    \hat{\mathbf{x}}_{t|t-1} = \sum_{i=0}^{2n} W_i^m\, f(\chi_{i}^{(t-1)}, \mathbf{u}_{t-1})
    \label{eq:ukf_predict_state}
\end{equation}
\begin{equation}
\begin{aligned}
    \mathbf{P}_{t|t-1} &= \sum_{i=0}^{2n} W_i^c 
    \big(f(\chi_{i}^{(t-1)}, \mathbf{u}_{t-1}) - \hat{\mathbf{x}}_{t|t-1}\big) \\
    &\quad \times \big(f(\chi_{i}^{(t-1)}, \mathbf{u}_{t-1}) - \hat{\mathbf{x}}_{t|t-1}\big)^\top
    + \mathbf{Q}_{t-1}
\end{aligned}
\label{eq:ukf_predict_uncer}
\end{equation}

where $\chi_{t-1}^i$ are the $2n\!+\!1$ sigma points generated from the previous state and covariance, $W_i^{m}$ and $W_i^{c}$ are the corresponding mean and covariance weights, and $\mathbf{Q}_{t-1}$ is the process noise covariance.

\textbf{Correction Step:}  
\begin{equation}
    \hat{\mathbf{x}}_{t} = \hat{\mathbf{x}}_{t|t-1} + \mathbf{K}_t\mathbf{y}_k
    \label{eq:ukf_update_state}
\end{equation}
\begin{equation}
    \mathbf{P}_{t} = \mathbf{P}_{t|t-1} - \mathbf{K}_t \mathbf{P}_{zz}\mathbf{K}_t^\top
    \label{eq:ukf_update_uncer}
\end{equation}
where $\mathbf{K}_t$ is the computed Kalman gain, $\mathbf{y}_k$ the computed measurement residual, and $\mathbf{P}_{zz}$ the computed innovation covariance.

\subsection{Bayesian Neural Kalman Filter}
\begin{figure}[t]
    \centering
    \includegraphics[width=\columnwidth]{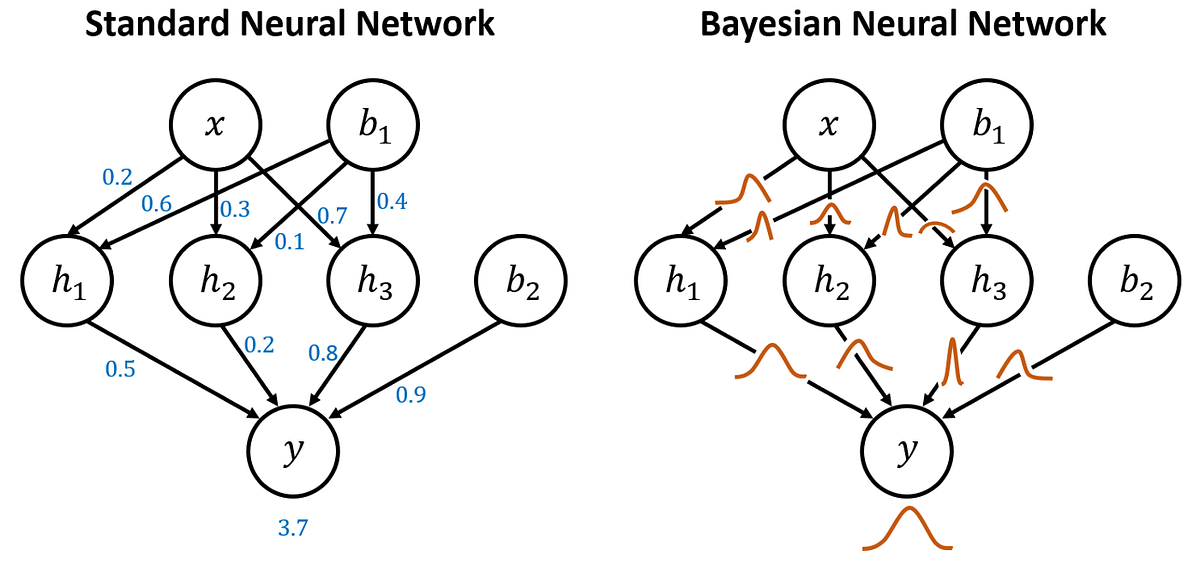}
    \caption{NN versus BNN \protect\cite{bnn_v_nn_graphic}}
    \label{fig:bnn_v_nn}
\end{figure}

Data-driven artificial neural networks have demonstrated superior performance over purely analytical linear solvers in fields such as signal processing \cite{osti_5470451} and biology \cite{ALMEIDA200272}. These models leverage historical data to predict future outcomes without requiring an explicit formulation of the underlying system dynamics. Bayesian Neural Networks (BNNs) extend this capability by incorporating uncertainty into their predictions, treating model weights as probability distributions rather than fixed values (Figure~\ref{fig:bnn_v_nn}). This approach enables the model to quantify its confidence in predictions by optimizing not only the standard Mean Squared Error (MSE) loss but also the Kullback-Leibler (KL) divergence (Equation~\ref{eq:MSE_Loss}), (Equation~\ref{eq:KLD_Loss}), (Equation~\ref{eq:total_loss}) \cite{bnn_uncer_explain}. The KL divergence term acts as a regularizer on the learned weight distributions, allowing principled uncertainty estimation. As a stochastic model, the BNN generates uncertainty estimates for its regression outputs through Monte Carlo (MC) sampling during inference. 

\begin{figure}[t] 
    \centering
    \includegraphics[width=\columnwidth]{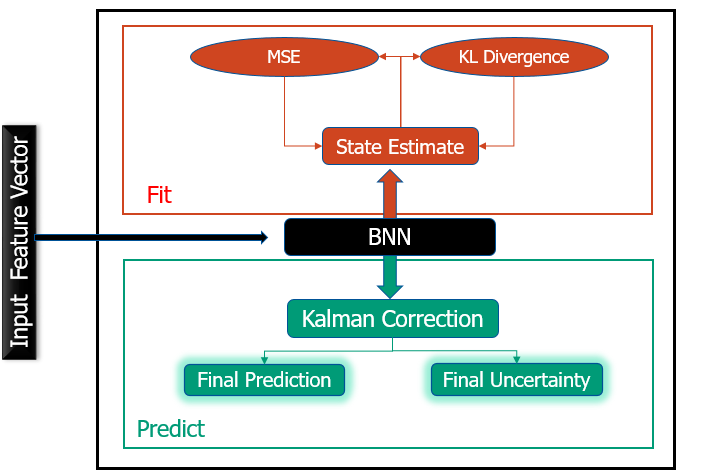}
    \caption{BNKF Process Flow}
    \label{fig:bnn_solution_diagram}
\end{figure}

We propose a \textit{BNKF}, which integrates the correction step of a standard EKF with a trained BNN that serves as a surrogate for the KF’s state-space model (Figure~\ref{fig:bnn_solution_diagram}). Ultimately, we leverage the statistical correction from the Kalman updater and apply it to a trained BNN's mean and uncertainty outputs (Equation~\ref{eq:bnn_est_state}), (Equation~\ref{eq:bnn_uncer_state}).

The trained Bayesian neural network (BNN) defines a mapping function 
$f: \mathbf{z}_t \mapsto (\hat{\mathbf{x}}_t, \mathbf{P}_t)$, which produces both a state estimate 
and its associated uncertainty from the feature vector $\mathbf{z}_t$. 
For clarity, we denote the two components of the output using $f_{\mu}$ 
for the mean (state estimate) and $f_{\Sigma}$ for the covariance (uncertainty). Here, predictive uncertainty arises from Monte Carlo sampling over variational weight distributions, providing an estimate of epistemic uncertainty that supplements measurement-driven correction in the subsequent Kalman update.

The position estimate at time $t$ is
\begin{equation}
    \hat{\mathbf{x}}_{t} 
    = f_{\mu}
    + \mathbf{K}_t\!\left(\mathbf{z}_t^{*} - \mathbf{H}_tf_{\mu}\right)
    \label{eq:bnn_est_state}
\end{equation}
\begin{equation}
    \mathbf{P}_{t} 
    = (\mathbf{I} - \mathbf{K}_t\mathbf{H}_t)\, f_{\Sigma}
    \label{eq:bnn_uncer_state}
\end{equation}
where $\mathbf{z}_t^{*}$ denotes the position measurement information derived from $\mathbf{z}_t$, and $\mathbf{H}_t$ is the observation matrix that maps the system state to the measurement space.
The analytical Kullback–Leibler (KL) divergence loss is
\begin{equation}
    L_{\text{KLD}} = -\frac{1}{2} \sum_{i=1}^{d} 
    \left(1 + \log(\sigma_i^2) - \mu_i^2 - \sigma_i^2 \right)
    \label{eq:KLD_Loss}
\end{equation}
where $d$ is the latent dimensionality, and $\mu_i$ and $\sigma_i^2$ denote the mean and variance of the learned Gaussian posterior for dimension $i$.
\vspace{-1ex}
The mean squared error (MSE) loss is
\begin{equation}
    L_{\text{MSE}} = \frac{1}{N} \sum_{i=1}^{N} (\mathbf{x}_i - \hat{\mathbf{x}}_i)^2
    \label{eq:MSE_Loss}
\end{equation}
where ${x}_i$ is the truth state and $N$ is the number of samples within the batch of a training epoch. The total loss function is then
\vspace{-1ex}
\begin{equation}
    L_T = L_{\text{MSE}} + L_{\text{KLD}}
    \label{eq:total_loss}
\end{equation}

\vspace{-1.0ex}

We believe training and integrating an intermediate BNN will help address EKF/UKF shortcomings in two ways: (1) Defining an appropriate process noise model for covariance prediction remains a challenge in traditional Kalman filtering \cite{process_noise_est}. Under varying sensor noise conditions, standalone EKF and UKF models cannot reliably adapt their noise covariance, which can degrade prediction performance \cite{ukf_noise_adaptability}. A trained BNN can instead adapt to this uncertainty by optimizing uncertainty estimates through KL divergence loss, improving covariance predictions. (2) In real-world scenarios, UAV motion is often driven more by control intent than by pure dynamics \cite{uav_unknown_intent}. Such intent-driven behavior is difficult for naive EKF/UKF models to capture during online tracking when the control policy is unknown. A trained BNN can instead learn these patterns from historical data while providing uncertainty-aware estimates.

In this approach, the BNN is trained to model the state prediction at the core, while the KF’s statistical correction refines the state prediction based on new measurements. The BNN leverages its learned understanding of nonlinear UAV control and associated uncertainty to provide a more accurate initial position and confidence estimate for the EKF correction step.

\subsection{Bayesian Neural Kalman Filter Ensemble}

We also evaluate an ensemble variant, \textit{BNKFe}, which decomposes state prediction into separate Cartesian components. Smaller BNNs are trained independently for $x$, $y$, and $z$, and their outputs are combined into a full state estimate and covariance matrix (Equation~\ref{eq:bnkfe_state}), (Equation~\ref{eq:bnkfe_uncer}). These outputs are then passed to the standard KF based corrector as we did with the \textit{BNKF} (Equation~\ref{eq:ekf_update_state}), (Equation~\ref{eq:ekf_update_uncer}). The goal of this ensemble design is to reduce individual model complexity while maintaining full-state predictive capability in a stable and robust manner. Ensemble machine learning architectures are generally intended as a technique to mitigate general model uncertainty in machine learning \cite{ensemble}. 

The intermediate mean state prediction is constructed by assembling the component-level predictions into a Cartesian state vector:

\begin{equation}
    \hat{\mathbf{x}}_{t} =
    \begin{bmatrix}
        \hat{x}_t \\[0.5ex]
        \hat{y}_t \\[0.5ex]
        \hat{z}_t
    \end{bmatrix}
    =
    \begin{bmatrix}
        f_{\mu_x}(\mathbf{z}_t) \\[0.5ex]
        f_{\mu_y}(\mathbf{z}_t) \\[0.5ex]
        f_{\mu_z}(\mathbf{z}_t)
    \end{bmatrix}
    \label{eq:bnkfe_state}
\end{equation}
where $f_{\mu_x}(\cdot)$, $f_{\mu_y}(\cdot)$, and $f_{\mu_z}(\cdot)$ denote the BNN outputs corresponding to the mean estimates along each Cartesian axis.

The associated uncertainty prediction is represented by a diagonal covariance matrix whose entries are the component-level variances predicted by the networks:
\begin{equation}
    \mathbf{P}_{t} =
    \begin{bmatrix}
        \big(f_{\Sigma_x}(\mathbf{z}_t)\big)^{2} & 0 & 0 \\
        0 & \big(f_{\Sigma_y}(\mathbf{z}_t)\big)^{2} & 0 \\
        0 & 0 & \big(f_{\Sigma_z}(\mathbf{z}_t)\big)^{2}
    \end{bmatrix}
    \label{eq:bnkfe_uncer}
\end{equation}
where $f_{\Sigma_x}(\cdot)$, $f_{\Sigma_y}(\cdot)$, and $f_{\Sigma_z}(\cdot)$ denote the BNN outputs corresponding to the standard deviations for each axis.

\begin{figure}[t] 
    \centering
    \includegraphics[width=\columnwidth]{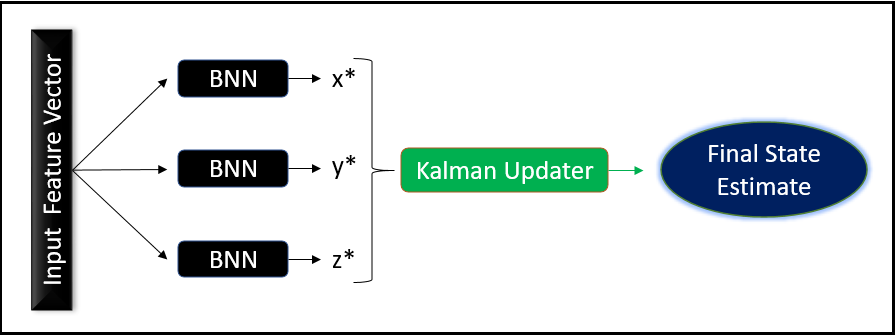}
    \caption{BNKFe Process Flow}
    \label{fig:bnkfe_solution_diagram}
\end{figure}

\section{Evaluation Procedure}\label{sec:evaluation}
In this section we outline the key evaluation metrics and experiments we utilized to gauge the performance of \textit{BNKF} relative to the benchmark algorithms.

\subsection{Metrics}
Metrics were chosen with the intention to provide a conjunctive evaluation of method accuracy and confidence. 
We evaluate performance using three metrics:  
\begin{enumerate}[label=(\arabic*), leftmargin=*, itemsep=0.5ex]

  \item \textbf{State Estimate Error [Euclidean Distance]:}
  \begin{equation}
    e_{t} = \sqrt{ ( \hat{x}_x - x_x )^2 + ( \hat{x}_y - x_y )^2 + ( \hat{x}_z - x_z )^2 }
    \label{eq:state_est_err}
  \end{equation}

  \item \textbf{State Estimate Uncertainty [Matrix Determinant]:}
  \begin{equation}
    \mathcal{V}_{t} = \det(\mathbf{P}_{t})
    \label{eq:state_est_uncer}
  \end{equation}

  \item \textbf{Truth Point Containment [Mahalanobis Distance]:}
  \begin{equation}
    D_t^2 = (\mathbf{x}_t - \hat{\mathbf{x}}_t)^\top \mathbf{P}_t^{-1} (\mathbf{x}_t - \hat{\mathbf{x}}_t)
    \label{eq:mahalanobis}
  \end{equation}
\end{enumerate}

For a consistent estimator in a three-dimensional state space, the expected value of $D_t^2$ equals the state dimension (i.e., $\mathrm{E}[D_t^2] \approx 3$). Deviations from this value indicate overconfident ($>3$) or conservative ($<3$) uncertainty estimates.

\subsection{Experiment Methodology}
We conducted three separate experiments corresponding to low, medium, and high sensor noise levels, computing our three primary evaluation metrics for each scenario (Equation~\ref{eq:state_est_err}), (Equation~\ref{eq:state_est_uncer}), (Equation~\ref{eq:mahalanobis}). Each noise level dataset was augmented with varying sampling rates, resulting in approximately 15,000 trajectories per test set. For the EKF and UKF, performance metrics were computed over all trajectories. For the BNN based models, we employed standard five-fold cross-validation and report the mean and standard deviation across folds. A standalone BNN (without the filter correction) was additionally included in the evaluation to isolate and demonstrate the performance gains attributable to the \textit{BNKF} framework. All neural networks in our experiments consisted of five Bayesian linear hidden layers with 64 neurons per layer, trained using the Adam optimizer \cite{kingma2017adammethodstochasticoptimization} and were Monte Carlo sampled for 100 realizations on inference. To support our Bayesian Neural Network, we utilized the torchbnn library \cite{torchbnn}.

\subsection{Limitations}
While the experimental methodology and results support the effectiveness of the \textit{BNKF} framework, several limitations/assumptions of the evaluation should be noted. 
(1) All experiments were conducted using synthetically generated UAV trajectories. Although the simulation environment is designed to be high fidelity, it does not fully capture the complexities of real-world data. 
(2) Training and evaluation assume measurements from a fixed sensor platform, which may limit applicability in scenarios involving moving or distributed sensors. 
(3) Sensor noise is modeled as Gaussian, whereas real-world sensing systems may exhibit more complex or non-Gaussian noise characteristics.

\section{Results}\label{sec:results}

\begin{figure}[t]
    \centering
    \includegraphics[
        width=\columnwidth,
        height=0.22\textheight,
        keepaspectratio
    ]{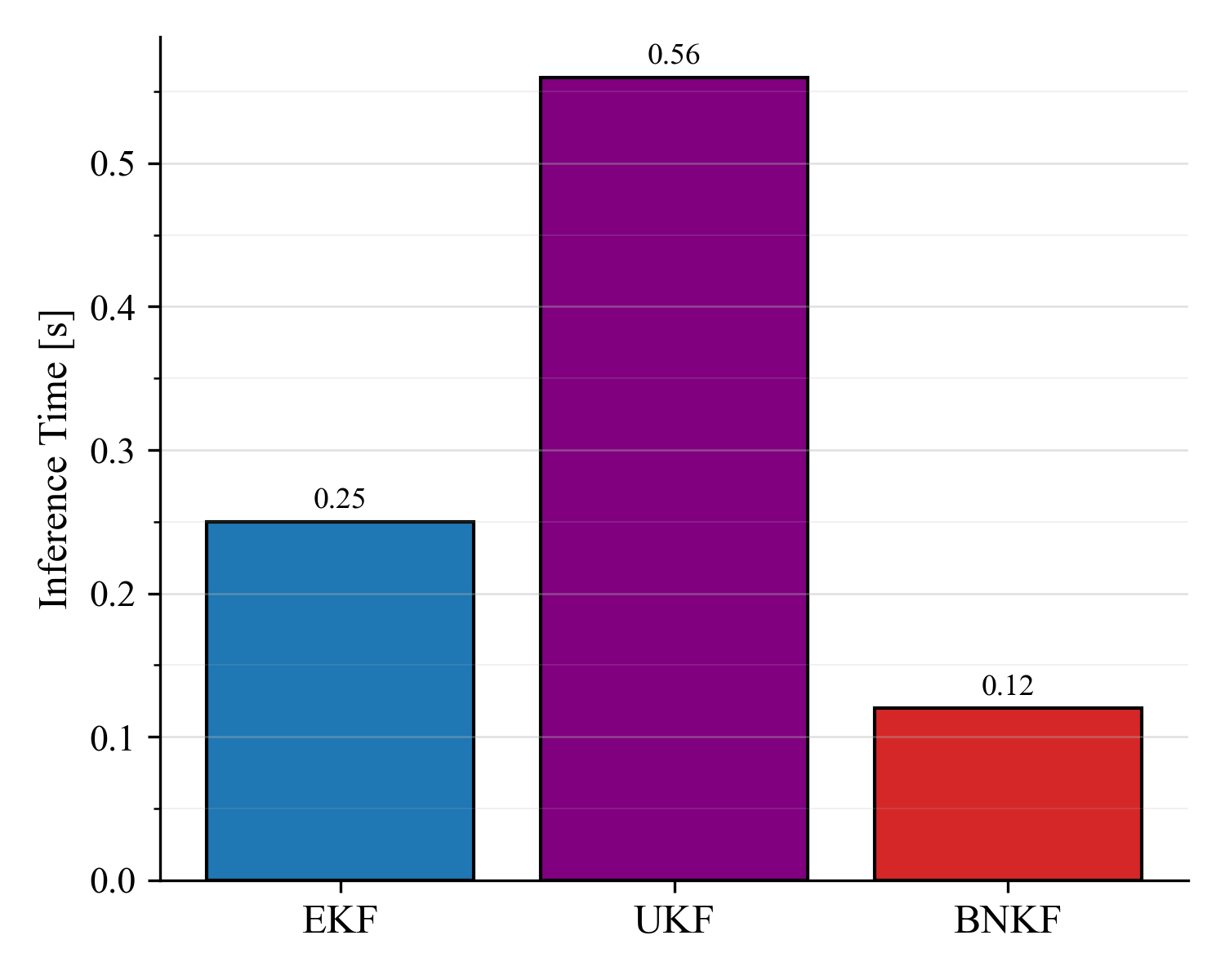}
    \caption{Single Trajectory Inference Time Comparison}
    \label{fig:run_time}
\end{figure}

\begin{figure}[t]
    \centering
    \includegraphics[
        width=\columnwidth,
        height=0.32\textheight,
        keepaspectratio
    ]{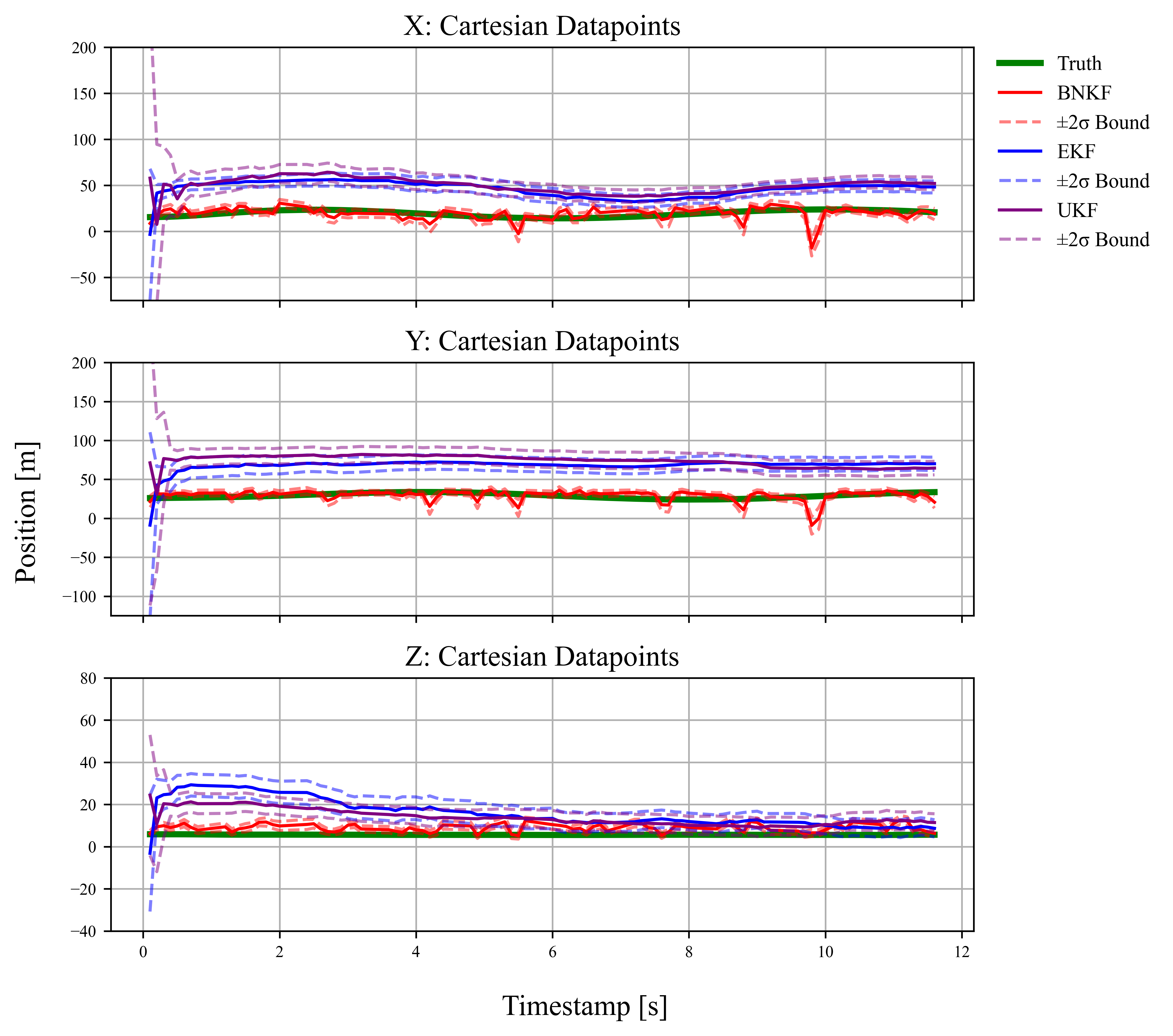}
    \caption{Example High Noise Trajectory Evaluation}
    \label{fig:traj_pred}
\end{figure}

\begin{figure}[t]
    \centering
    \includegraphics[
        width=\columnwidth,
        height=0.45\textheight,
        keepaspectratio
    ]{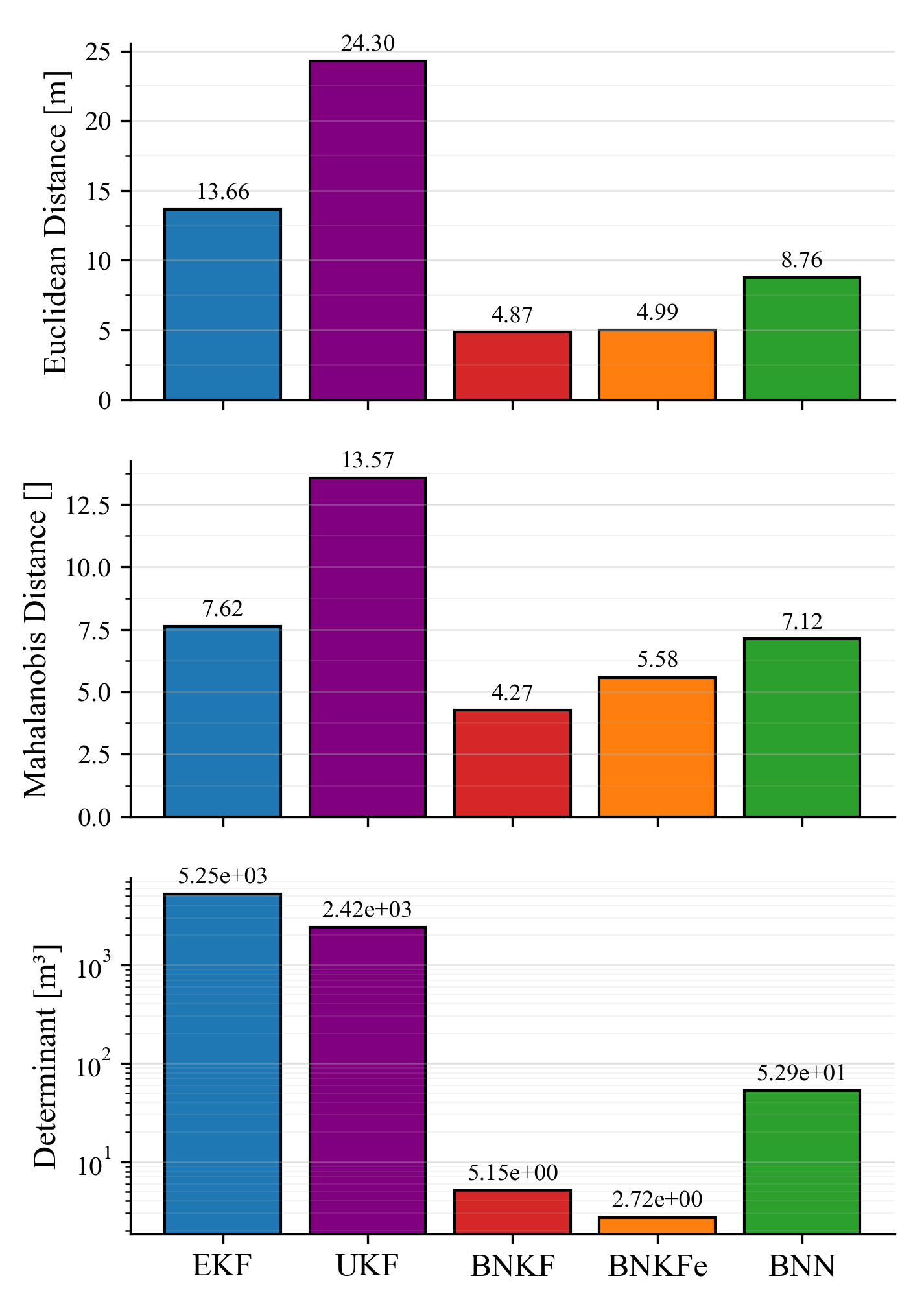}
    \caption{Noise Level Averaged Method Performance}
    \label{fig:noise_level_averaged}
\end{figure}

\begin{figure*} 
    \centering
    \begin{subfigure}{0.31\textwidth}
        \centering
        \includegraphics[width=\linewidth]{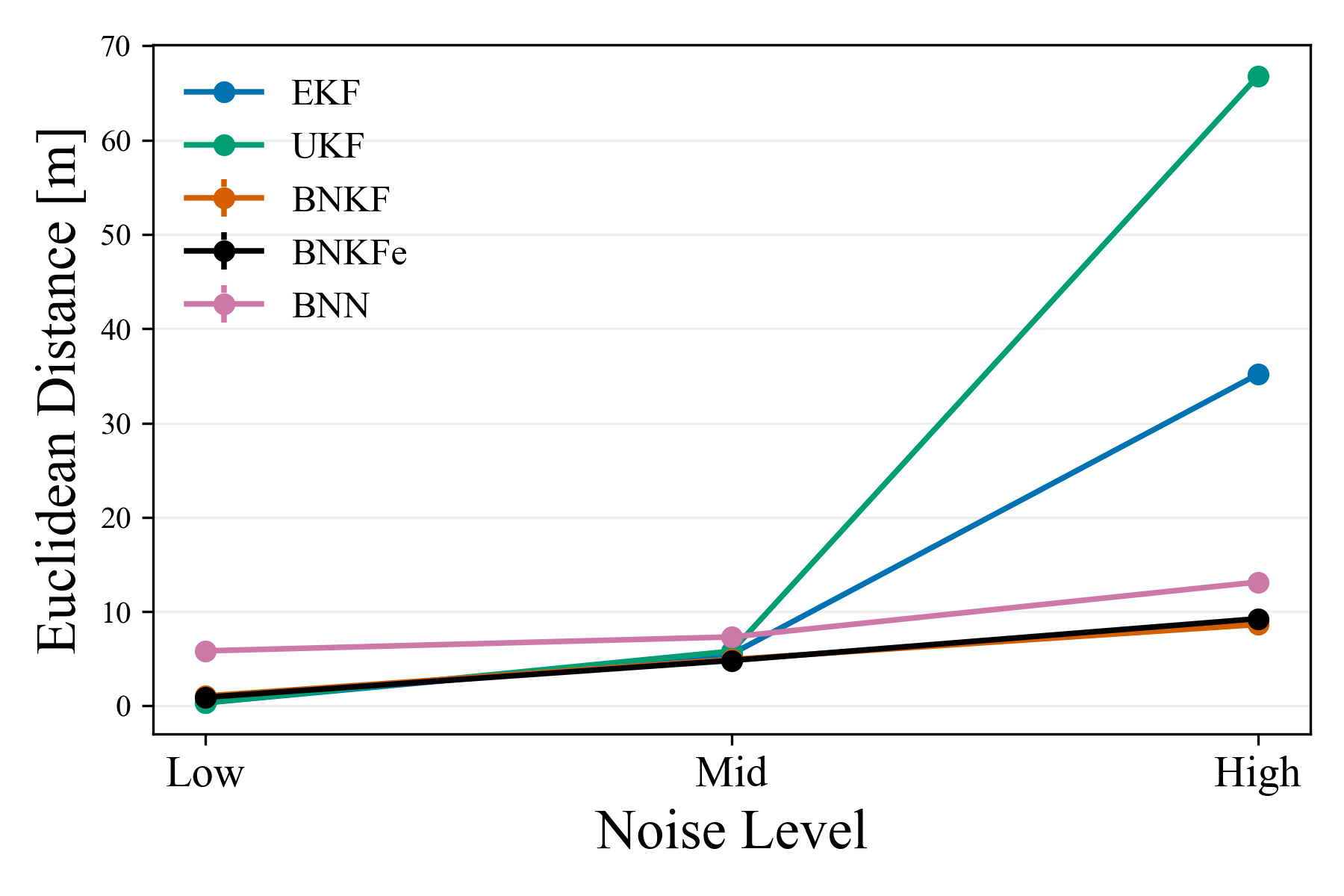}
        \caption{ED Comparison}
        \label{fig:ed_comp}
    \end{subfigure}
    \hfill
    \begin{subfigure}{0.31\textwidth}
        \centering
        \includegraphics[width=\linewidth]{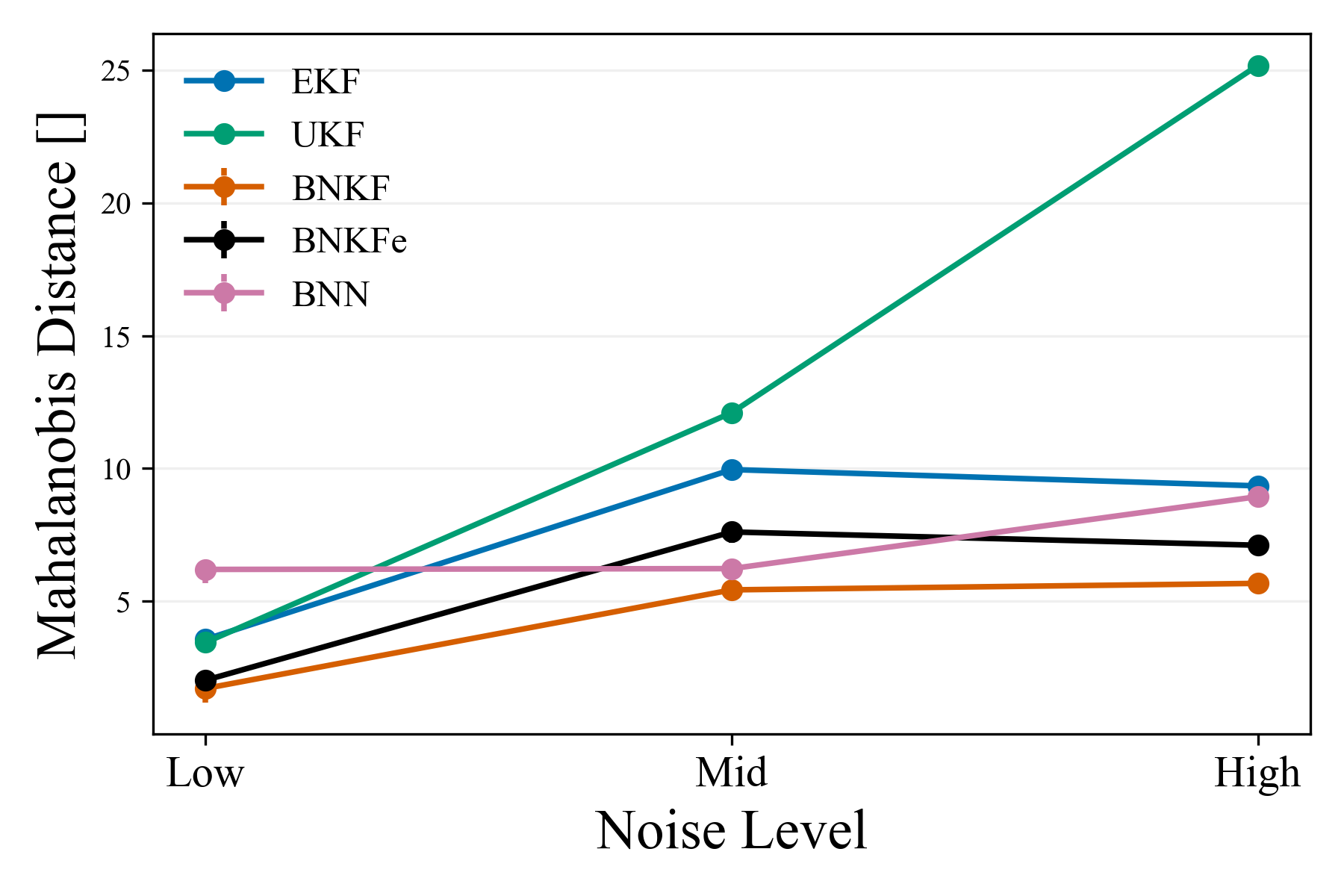}
        \caption{MD Comparison}
        \label{fig:md_comp}
    \end{subfigure}
    \hfill
    \begin{subfigure}{0.31\textwidth}
        \centering
        \includegraphics[width=\linewidth]{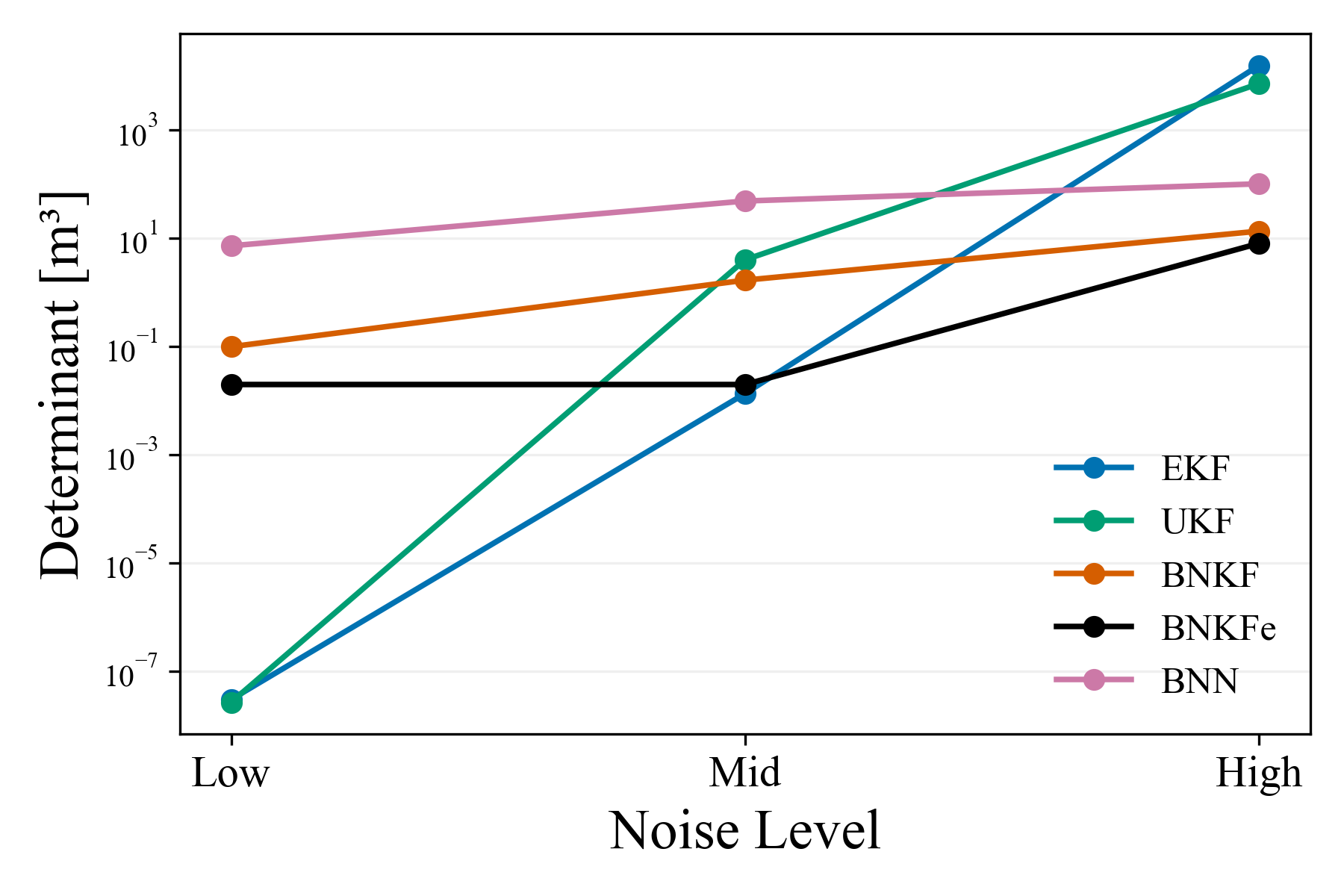}
        \caption{Determinant Comparison}
        \label{fig:det_comp}
    \end{subfigure}
    \caption{Method Comparison across Metrics}
    \label{fig:method_comp_p_metric}
\end{figure*}

We present the results from the three sensor noise experiments along with their corresponding evaluations. The single-trajectory plots highlight how differences between the filters manifest at the scale of individual runs, while the aggregated plots capture broad patterns across hundreds of trajectories and multiple noise regimes. Table~\ref{tab:written_metrics_results} provides a summary of each method performance across all metrics broken down by simulated sensor measurement noise. Figure~\ref{fig:run_time} show each method's inference runtime on a full single example trajectory utilizing a  CPU. Figure~\ref{fig:traj_pred} shows a comparative example of the models inference evaluated on a high noise - 75 percent sample rated trajectory. Figure~\ref{fig:noise_level_averaged} visualizes each method's performance across each evaluation metric averaged across all tested noise levels for the entire test dataset. Figure~\ref{fig:method_comp_p_metric} breaks this down and showcases method performance as a function of the different sensing noise levels for the entire test dataset. Altogether, we observe that the \textit{BNKF} generally outperforms the benchmark KF's across multiple metrics within higher noise levels. 

\begin{table*} 
\renewcommand{\arraystretch}{1.3} 
\centering
\tiny
\resizebox{\textwidth}{!}{%
\begin{tabular}{|c|c|c|c|c|c|c|c|c|c|}
    \hline
    \textbf{Models} & \multicolumn{3}{c|}{\textbf{Low Noise}} 
                    & \multicolumn{3}{c|}{\textbf{Mid Noise}} 
                    & \multicolumn{3}{c|}{\textbf{High Noise}} \\ \hline
    & \textbf{ED [m]} & \textbf{MD []} & \textbf{Det [m$^3$]} 
    & \textbf{ED [m]} & \textbf{MD []} & \textbf{Det [m$^3$]} 
    & \textbf{ED [m]} & \textbf{MD []} & \textbf{Det [m$^3$]} \\ \hline

    EKF   & 0.33 $\pm$ 0 & 3.57 $\pm$ 0 & 3.02e-8 $\pm$ 0 
          & 5.49 $\pm$ 0 & 9.96 $\pm$ 0 & 0.03 $\pm$ 0 
          & 35.16 $\pm$ 0 & 9.35 $\pm$ 0 & 15740 $\pm$ 0 \\ \hline

    UKF   & 0.33 $\pm$ 0 & 3.45 $\pm$ 0 & 2.68e-8 $\pm$ 0
          & 5.80 $\pm$ 0 & 12.10 $\pm$ 0 & 4.03 $\pm$ 0
          & 66.78 $\pm$ 0 & 25.17 $\pm$ 0 & 7262 $\pm$ 0 \\ \hline

    BNN   & 5.83 $\pm$ 0.32 & 6.20 $\pm$ 0.52 & 7.31 $\pm$ 7.31
          & 7.31 $\pm$ 0.42 & 6.23 $\pm$ 0.32 & 49.22 $\pm$ 12.19
          & 13.13 $\pm$ 0.33 & 8.94 $\pm$ 0.19 & 102.26 $\pm$ 24.82 \\ \hline

    \textit{BNKF}   & 1.05 $\pm$ 0.52 & \textbf{1.72 $\pm$ 0.52} & 0.09 $\pm$ 0.13
          & 4.92 $\pm$ 0.03 & \textbf{5.43 $\pm$ 0.09} & 1.70 $\pm$ 0.13
          & \textbf{8.63 $\pm$ 0.18} & \textbf{5.68 $\pm$ 0.15} & 13.66 $\pm$ 2.16 \\ \hline

    \textit{BNKFe}  & 0.91 $\pm$ 0.05 & 2.03 $\pm$ 0.07 & 0.02 $\pm$ 0.00
          & \textbf{4.81 $\pm$ 0.08} & 7.61 $\pm$ 0.42 & \textbf{0.02 $\pm$ 0.02}
          & 9.24 $\pm$ 0.13 & 7.11 $\pm$ 0.36 & \textbf{8.11 $\pm$ 1.90} \\ \hline

\end{tabular}%
}
\caption{Method Performance Metric Numerical Comparison (Areas of performance gain \textbf{bolded})}
\label{tab:written_metrics_results}
\end{table*}

\section{Discussion}
Overall, the \textit{BNKF} outperforms its EKF and UKF counterpart when metrics are averaged across all scenarios (Figure~\ref{fig:noise_level_averaged}). We additionally note that its advantage becomes pronounced in mid to high noise level datasets as sensor noise increases  (Figure~\ref{fig:method_comp_p_metric}). As shown in Table~\ref{tab:written_metrics_results}, the \textit{BNKF} and its ensemble variant achieve superior performance on an increasing number of metrics by an increased margin as noise levels rise. Notably, the \textit{BNKF} maintains the lowest average Mahalanobis Distance (MD) across all noise settings; however, this may partly reflect its tendency to produce larger uncertainty estimates in low to mid noise regimes since its determinant remains higher. Larger uncertainty “bubbles” increase the likelihood of containing the true state statistically, potentially inflating containment-related metrics.

More importantly, the gap between \textit{BNKF} and the UKF/EKF widens significantly at higher noise levels. We observe an inflation in uncertainty projection from the standard KF approaches with degraded accuracy as well. This is likely due to the fact that the UKF and EKF are unable to dynamically adapt their process noise covariance models in response to the increasing measurement noise, showing a limitation in robustness. In contrast, the \textit{BNKF} benefits from the uncertainty projection optimization during BNN training, enabling it to maintain relatively low determinant and MD values even under high noise. This suggests improved calibration of its uncertainty estimates, contributing to more reliable and accurate state estimation. Figure~\ref{fig:traj_pred} highlights a representative high-noise trajectory where 25 percent of observations were removed. Here, the \textit{BNKF} demonstrates a clear advantage, with an average Euclidean Distance (ED) error of 9 meters compared to 30+ meters for the UKF/EKF—a substantial improvement in state estimation accuracy. Here we also note that the projected uncertainty bounds the error consistently among all three methods, implying that the \textit{BNKF} maintains a well calibrated uncertainty. While this example is illustrative for a single trajectory, the comprehensive results in Table~\ref{tab:written_metrics_results} confirm the consistency of the \textit{BNKF}’s superior performance across the entire trajectory test set.

We note that the ensembled variant \textit{BNKFe} can offer a reduced amount of uncertainty in its state estimates relative to the \textit{BNKF} while maintaining similar accuracy. This outcome is consistent with expectations, as ensemble models generally reduce variability in estimates (Figure~\ref{fig:noise_level_averaged}), (Table~\ref{tab:written_metrics_results}). To assess the value of the post-inference Kalman correction, we include a standalone BNN as a control model. As shown in Figures~\ref{fig:noise_level_averaged} and Table~\ref{tab:written_metrics_results}, the correction step leads to consistent improvements, with both \textit{BNKF} and \textit{BNKFe} surpassing the standalone BNN across all metrics.

Lastly, Figure~\ref{fig:run_time} shows that \textit{BNKF} maintains a lower computational cost for trajectory prediction compared to the EKF and UKF, addressing a common concern with neural network–driven algorithms. However, this comparison assumes the BNKF has already been trained and reflects only inference of the prediction algorithm on a CPU.

\section{CONCLUSIONS}

Conclusively, this paper introduces a novel approach, \textit{BNKF}, to address the UAV state estimation problem in the presence of noisy and sparse sensor measurements. Such diverse sensing conditions are more prevalent than commonly assumed and can arise due to unexpected environmental factors, adversarial jamming, or manufacturing defects. We demonstrate that \textit{BNKF} can outperform a baseline EKF and UKF across various sensor noise and observation availability scenarios in position estimation accuracy, uncertainty quantification, and Mahalanobis Distance minimization. We posit that the \textit{BNKF}’s principal advantage over traditional Kalman Filtering techniques emerges under high noise and sparse sampling scenarios, where standard models struggle to adapt dynamically. Ensembling the \textit{BNKF} reduces overall model uncertainty across all noise levels with an additional accuracy improvement for higher noise levels. We also prove that the coupled filter correction step provides utility in refining the underlying BNN's initial inference. Finally, we demonstrate that the \textit{BNKF} is still able to predict with low computational expense compared to traditional methods. Beyond performance improvements, this study highlights the promising potential of hybridizing data-driven artificial intelligence with principled computational frameworks to create a more robust and reliable state prediction system. This hybrid approach opens new avenues for robust estimation in complex, uncertain environments, offering valuable insights for future research and practice. 

\addtolength{\textheight}{-4cm}   




\def\UrlBreaks{\do\/\do-}

\noindent\textbf{Code Availability:}
All source code used in this study is available at our GitHub repository:\url{https://github.com/agupt126/NeuralAugmentedKalmanFiltering_BNKF/}.
\section*{ACKNOWLEDGMENT}

We acknowledge the support of the Johns Hopkins University Engineering for Professionals (JHUEP) program. We are grateful to our colleagues Aniket Goel and Seth Graham at the Johns Hopkins University Applied Physics Laboratory (JHUAPL) for their valuable consulting and insights throughout this work. We also acknowledge the assistance of scite.ai in supporting the literature review process and OpenAI's ChatGPT in refining the clarity of the manuscript.


\end{document}